\documentclass{article} 
\usepackage[final]{colm2026_conference}

\usepackage{microtype}
\usepackage{hyperref}
\usepackage{url}
\usepackage{booktabs}
\usepackage[pass]{geometry}
\usepackage{graphicx}
\usepackage{amsmath} 
\usepackage{multirow}
\usepackage{pgfplots}

\usepackage{lineno}

\definecolor{darkblue}{rgb}{0, 0, 0.5}
\hypersetup{colorlinks=true, citecolor=darkblue, linkcolor=darkblue, urlcolor=darkblue}

\title{TGIF: Text-Guided Layer Fusion Mitigates Hallucination in Multimodal LLMs}

\author{
Chenchen Lin\textsuperscript{1},
Sanbao Su\textsuperscript{1}\thanks{Current affiliation: Google.},
Rachel Luo\textsuperscript{2},
Yuxiao Chen\textsuperscript{2},
Yan Wang\textsuperscript{2},\\
~\textbf{Marco Pavone}\textsuperscript{2,3}\thanks{Corresponding author.},
~\textbf{Fei Miao}\textsuperscript{1}\thanks{Current affiliation: Zhejiang University. Corresponding author.}
\\[0.5em]
\textsuperscript{1}University of Connecticut \qquad
\textsuperscript{2}NVIDIA \qquad
\textsuperscript{3}Stanford University
\\[0.4em]
\texttt{\{chenchen.lin, sanbao.su, fei.miao\}@uconn.edu}
\\
\texttt{\{raluo, yuxiaoc, yanwan, mpavone\}@nvidia.com}
}
\begin{document}

\ifcolmsubmission
\linenumbers
\fi

\maketitle

\begin{abstract}
Multimodal large language models (MLLMs) often generate confident but visually ungrounded responses, a phenomenon known as hallucination.
While prior mitigation methods primarily operate at decoding time or rely on additional training, they typically expose the language model to a fixed, late-layer visual representation, overlooking the rich hierarchy encoded by vision transformers.
We show that hallucination behavior is strongly influenced by the depth of visual features provided to the LLM, and that no single layer is optimal across queries.
To address this, we propose Text-Guided Inter-layer Fusion (TGIF), a lightweight architectural module that dynamically reweights visual features across transformer layers based on the input text, without modifying the vision encoder or increasing the token budget.
Experiments on hallucination, OCR, and general VQA benchmarks demonstrate that TGIF substantially improves visual grounding and hallucination robustness while preserving strong overall reasoning performance. \textbf{Code:} \href{https://github.com/Linchenchen/TGIF}{github.com/Linchenchen/TGIF}.
\end{abstract}
\section{Introduction}
Multimodal large language models (MLLMs) combine the reasoning capabilities of large language models (LLMs) with pretrained vision encoders, achieving impressive results in diverse visual tasks~\cite{llava,li2023blip,instructblip}. However, MLLMs face  persistent hallucination challenge, generating confident but ungrounded content that is plausible under language priors yet inconsistent with the image~\cite{liu2024surveyhallucinationlargevisionlanguage}. This is particularly pronounced in detail-oriented tasks where high-level semantic features alone are insufficient for precise grounding. Most existing hallucination mitigation methods intervene after training or at inference time~\cite{gunjal2024detectingpreventinghallucinationslarge,yu2024rlhfvtrustworthymllmsbehavior,leng2023mitigatingobjecthallucinationslarge, liu2024reducinghallucinationsvisionlanguagemodels, huang2024opera, tang2025seeingfarclearlymitigating, chen2025perturbollavareducingmultimodalhallucinations}.
While effective, these approaches primarily operate on the text generation process and leave the underlying visual representations unchanged.  

\begin{figure}[t]
    \centering
    \includegraphics[width=0.9\linewidth,trim={3.5cm 2cm 4cm 2cm},clip]{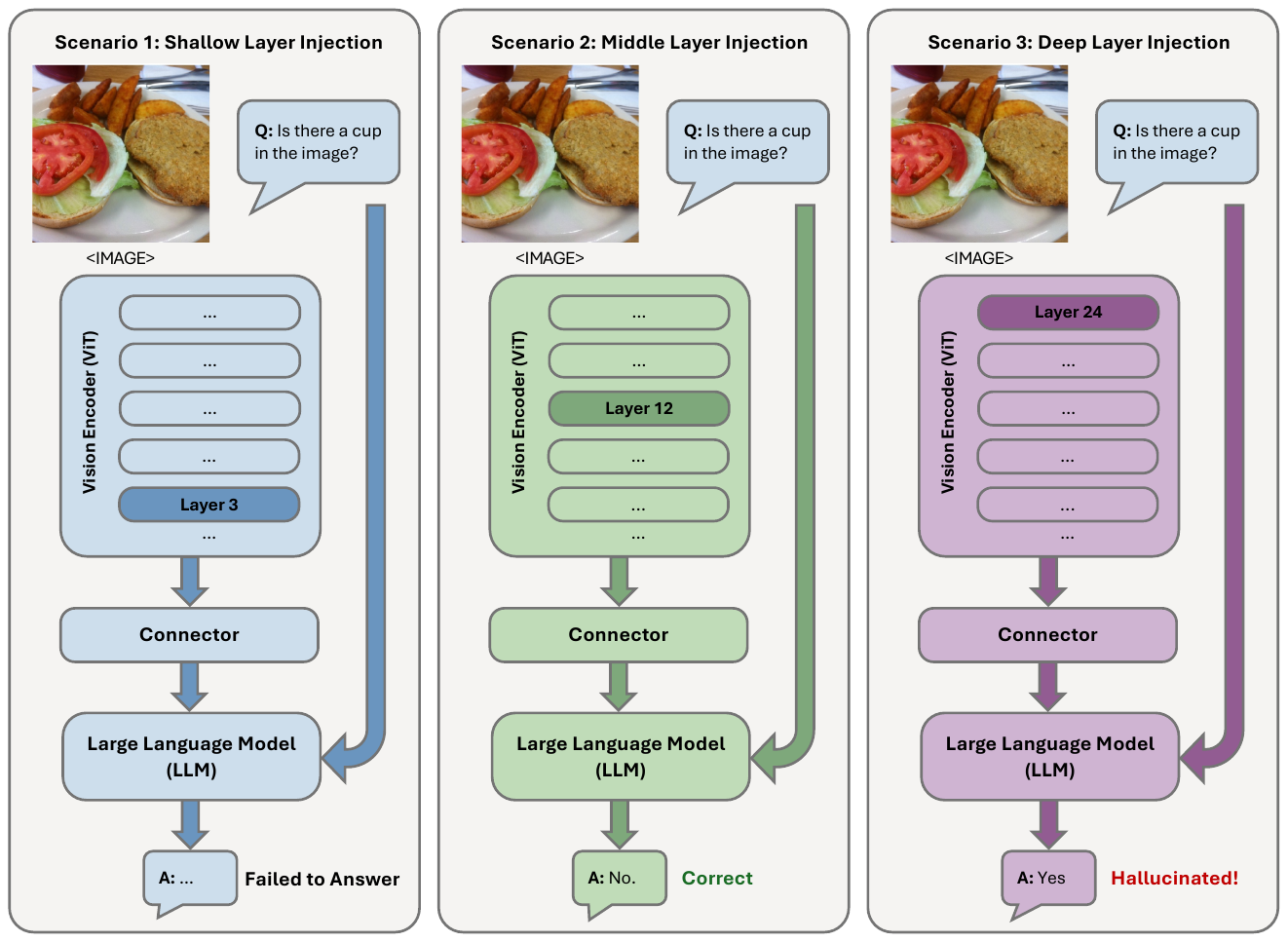}
    \caption{\textbf{Impact of ViT Layer Injection on Hallucination Behavior.} }
    \label{fig:vit-layer-injection}
    \vskip -0.2in
\end{figure}

A key but underexplored factor is the depth of visual representations exposed to the language model.
Vision transformers such as CLIP encode a hierarchy of visual abstractions across layers, ranging from low-level spatial cues to high-level semantic concepts.
However, most MLLMs expose only a single, fixed (typically late) vision layer to the LLM via an MLP projector.
Figure~\ref{fig:vit-layer-injection} presents a representative example in which the baseline LLaVA model hallucinates, demonstrating that injecting visual features from different depths, with all other components held fixed, results in qualitatively different behaviors ranging from under-recognition to hallucination.

Recent works begin to highlight the role of visual encoder layers in MLLMs, suggesting that shallow and middle layers exhibit unique strengths over deep layers in certain sub-tasks~\cite{chen2025multimodallanguagemodelsbetter}.
VaLiD~\cite{liu2024reducinghallucinationsvisionlanguagemodels} and
LayerCD~\cite{chen2025perturbollavareducingmultimodalhallucinations} collectively suggest that hallucination behavior is sensitive to the choice of visual feature depth. To the best of our knowledge, however, no prior work has addressed this issue by explicitly modifying the multimodal architecture to dynamically control which visual representations are exposed to the LLM.

\begin{figure*}[t]
    \centering
    \includegraphics[width=0.9\textwidth,trim={0.5cm 8.7cm 10cm 2.7cm},clip]{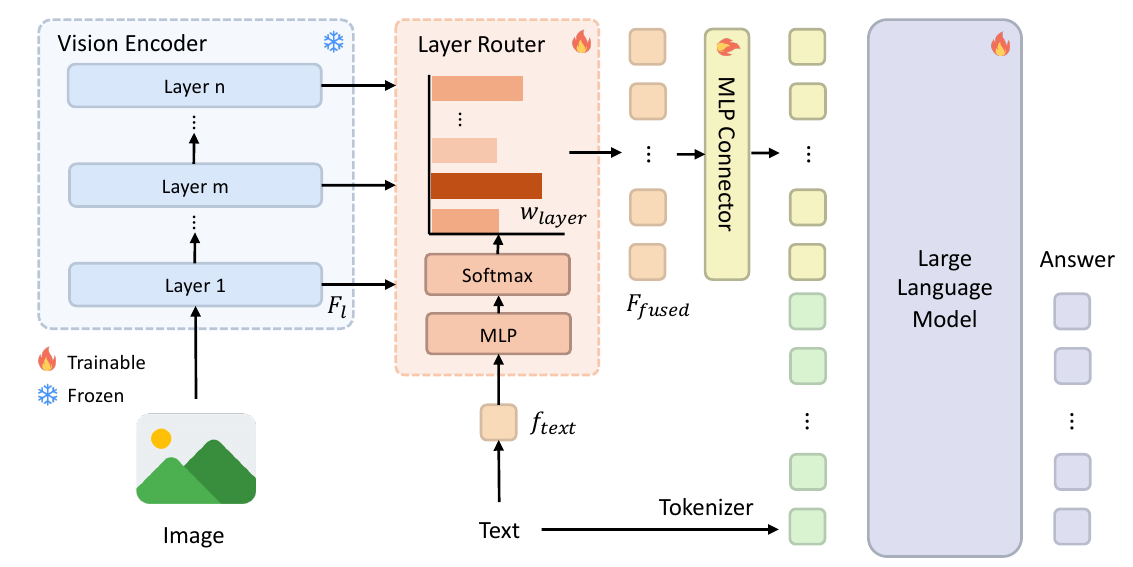}
    \caption{\textbf{Overview of the proposed Text-Guided Inter-Layer Fusion (TGIF) framework.}
   TGIF dynamically integrates multi-layer visual features $\{F_l\}$ from a frozen ViT based on the textual query. 
A Layer Router processes the text embedding $f_{\text{text}}$ to produce a soft distribution $w_{\text{layer}}$ via an MLP and softmax. 
These weights fuse hierarchical representations into $F_{\text{fused}}$, which is then projected and concatenated with text tokens for LLM reasoning.}
\label{fig:2}
\vskip -0.1in
\end{figure*}

Motivated by this observation, we hypothesize that hallucination in MLLMs is strongly influenced by how visual features are selected and exposed to the language model, and that adapting this selection based on the input prompt can improve visual grounding.
We propose \textbf{TGIF} (\textbf{T}ext-\textbf{G}uided \textbf{I}nter-layer \textbf{F}usion), a lightweight architectural module that dynamically reweights visual features across all layers of a frozen vision encoder according to the textual query (Fig.~\ref{fig:2}).
This design preserves the pretrained backbone and token budget while enabling prompt-dependent control over visual abstraction.

We instantiate TGIF on top of LLaVA-1.5 and evaluate it on hallucination, OCR, and general VQA benchmarks.
TGIF consistently improves hallucination robustness and fine-grained visual perception while maintaining competitive performance on general reasoning tasks, demonstrating that better visual grounding can be achieved through dynamic representation-level control rather than post hoc decoding intervention.
Our main contributions are three-fold:
\begin{itemize}
    \item We identify a limitation of current MLLMs: visual tokens are typically drawn from a single late-layer representation, which is poorly suited for detail-sensitive grounding and exacerbates hallucination under strong language priors.
    \item We propose \textbf{TGIF}, a text-guided inter-layer fusion module that dynamically reweights CLIP layers per query while remaining parameter- and token-efficient. 
    \item We show that TGIF improves hallucination robustness and fine-grained visual perception on POPE, HallusionBench, OCRBench, and TextVQA, while maintaining competitive performance on general reasoning benchmarks.
\end{itemize}
\section{Related Work}
\subsection{Multimodal Large Language Models}
Multimodal large language models (MLLMs) typically combine a frozen vision encoder with a language model through lightweight connectors~\cite{llava, li2023blip, instructblip}. Early designs adopt simple MLP projectors~\cite{llava}, while later approaches introduce query-based extraction mechanisms~\cite{instructblip}. Despite these architectural differences, most MLLMs rely on a \emph{fixed} visual interface, commonly projecting features from a single late layer of the vision encoder. This design implicitly assumes that a single level of visual abstraction is sufficient across all queries, leaving limited flexibility to adapt to diverse grounding requirements.
\subsection{Hallucination and Representation Depth}
Hallucination in MLLMs, where models generate visually ungrounded content, has been widely studied~\cite{rohrbach2019objecthallucinationimagecaptioning}. Existing mitigation strategies largely operate at the decoding or training level, including RLHF-based alignment~\cite{gunjal2024detectingpreventinghallucinationslarge, yu2024rlhfvtrustworthymllmsbehavior} and decoding-time interventions such as VCD, OPERA,  FarSight and SHIELD~\cite{leng2023mitigatingobjecthallucinationslarge, huang2024opera, tang2025seeingfarclearlymitigating, huang2026shield}. Both LayerCD~\cite{tong2025mitigatinghallucinationmultimodalllms} and VaLiD~\cite{wang2025validmitigatinghallucinationlarge} investigate the role of vision encoder layers by coupling hierarchical feature exploration with contrastive decoding strategies. However, recent work suggests that such improvements may not always correspond to genuine grounding gains~\cite{yin2025mirageperformancegainscontrastive}, motivating investigation into the representation stage.
\subsection{Multi-Layer Feature Integration}
Transformer-based vision encoders exhibit a hierarchical structure, where shallow and intermediate layers preserve spatial details while deeper layers capture semantic abstractions~\cite{clip,gandelsman2024interpretingclipsimagerepresentation}. Prior work has explored integrating multiple layers to enrich visual representations. For example, DenseConnector~\cite{yao2024denseconnectormllms} concatenates features from different depths, while MMFuser~\cite{cao2024mmfusermultimodalmultilayerfeature} retrieves fine-grained information using deep-layer queries. More recent analyses further suggest that exposing multiple layers alone is insufficient, and that the effectiveness of different layers depends on the task and query~\cite{lin2025multilayervisualfeaturefusion}. Building on these insights, TGIF performs text-guided inter-layer fusion, dynamically reweighting visual features according to the input query to improve grounding and reduce hallucination.
\section{Method}
Our approach is situated within a standard Vision Language Model (VLM) architecture, which comprises a vision encoder (e.g., CLIP ViT~\cite{clip}), a language model (e.g., Vicuna~\cite{vicuna}), and a multimodal projector that maps visual features into the language model's embedding space. Our proposed text-guided layer selection module is designed as a core component of this multimodal projector. It takes the full stack of hidden states from the all vision encoder layers as input, along with features derived from the text prompt, to produce a dynamically tailored visual representation for the LLM.
\subsection{Text-Guided Layer Selection}
As established in prior work~\cite{chen2025multimodallanguagemodelsbetter}, different CLIP layers capture distinct semantic information, with shallow layers encoding textures and spatial details while deeper layers align more with global semantics. This suggests that the layers can be viewed as a pool of specialized ``experts''. Inspired by the Mixture-of-Experts (MoE) paradigm, we treat each layer as an expert and propose a dynamic layer selection framework, called TGIF (Text-Guided Inter-layer Fusion). In this framework, a text-guided ``router'' learns to generate weights to select and fuse the most relevant layer experts for a given task. We explore two architectures for the TGIF router, one with text-only input and other with multimodal input. The model architecture and framework is shown in Fig~\ref{fig:2}. This design allows adaptive, query-conditioned fusion of depth-wise visual cues, improving both grounding and fine-grained detail understanding.
\subsubsection{Text-Guided MLP Router}
Our baseline approach uses a lightweight MLP-based router to predict layer importance scores based solely on the textual prompt. This router learns a direct mapping from the question's semantics to the relevance of different vision layers. Let $\mathbf{f}_{\text{text}} \in \mathbb{R}^{D_t}$ be the pooled text embedding from the LLM, $\{\mathbf{F}_{l} \in \mathbb{R}^{P \times D_v}\}_{l=1}^L$ be the set of patch-level visual features from all $L$ layers of the vision encoder. The MLP selector first predicts unnormalized logits for each layer:
\begin{equation}
\mathbf{z} = \text{MLP}(\mathbf{f}_{\text{text}}) \in \mathbb{R}^{L}.
\end{equation}
These logits are then transformed into a probability distribution using the softmax function to represent the layer weights:
\begin{equation}
\boldsymbol{w} = \text{softmax}(\mathbf{z}) \in \mathbb{R}^{L}.
\end{equation}
The final fused visual representation is computed as a weighted sum of all layer features, where the weights are broadcast across the patch and feature dimensions:
\begin{equation}
\mathbf{F}_{\text{fused}} = \sum_{l=1}^{L} w_l \cdot \mathbf{F}_{l} \in \mathbb{R}^{P \times D_v}.
\end{equation}
\subsubsection{Multimodal MLP Router}
To address cases in pretraining where the text prompt is generic (e.g., ``Describe the image''), we enhance the router with visual context. This multimodal approach allows the selection to be conditioned on both the question and the image content. 
We first extract a global image representation, $\mathbf{f}_{\text{image}} \in \mathbb{R}^{D_v}$, by taking the [CLS] token from the penultimate layer of the vision encoder. The text and image features are then projected to a common dimension $D_p$ and concatenated:
\begin{equation}
\mathbf{f}_{\text{multi}} = [\mathbf{f}_{\text{text}}\mathbf{W}_{t}, \mathbf{f}_{\text{image}}\mathbf{W}_{v}] \in \mathbb{R}^{2D_p},
\end{equation}
where $\mathbf{W}_{t} \in \mathbb{R}^{D_t \times D_p}$ and $\mathbf{W}_{v} \in \mathbb{R}^{D_v \times D_p}$ are learnable projection matrices. This combined multimodal feature vector is then used by the MLP to predict the layer weights, following the same process as in Equations 1-3. A schematic comparison between the text-only and multimodal routers is provided in Appendix~\ref{sec:router_variants} (Fig.~\ref{fig:router_variants}).
\subsection{Load Balancing Loss}
\label{sec:load_balance}
A common challenge when training MoE-style routers is the tendency for the router to converge to a state where it consistently selects the same few ``safe'' experts (in our case, layers), leading to ``expert starvation''~\cite{shazeer2017outrageouslylargeneuralnetworks}. To mitigate this and encourage the router to utilize a more diverse set of layers, we incorporate a modified auxiliary load balancing loss into our total training objective.
For our soft-selection routers, we use an entropy-based loss. Let $\boldsymbol{w}_b \in \mathbb{R}^L$ be the layer weights for the $b$-th sample in a batch of size $B$. We first compute the average weight for each layer across the batch:
\begin{equation}
\bar{\boldsymbol{w}} = \frac{1}{B} \sum_{b=1}^B \boldsymbol{w}_b \in \mathbb{R}^L.
\end{equation}
The auxiliary loss is then formulated to maximize the entropy of this average distribution, which encourages a more uniform usage of layers:
\begin{equation}
\mathcal{L}_{\text{aux}} = \lambda \sum_{l=1}^L \bar{w}_l \log(\bar{w}_l + \epsilon),
\end{equation}
where $\lambda$ is a hyperparameter controlling the strength of the loss and $\epsilon$ is a small constant for numerical stability. This auxiliary loss is added to the main VLM loss during training.
Because the nature of textual prompts differs between pretraining and instruction tuning dataset, we apply different $\lambda$ values across stages. 
During pretraining, the router often receives generic prompts (e.g., “Describe the image”), which provide limited textual guidance. 
We thus strengthen the visual signal by applying a slightly larger $\lambda$ to encourage exploration of multiple layers.  
During fine-tuning, prompts are task-oriented and semantically rich (e.g., “What number is written on the sign?”). 
So we reduce $\lambda$ to allow the router to focus on discriminative, text-conditioned layer selection.
\section{Experiments}
\subsection{Experimental Setting}
We implement our proposed framework, TGIF, on top of the publicly available LLaVA-1.5~\cite{li2023llava} codebase. To ensure a fair comparison, our primary experiments maintain consistency with LLaVA-1.5 by employing CLIP-ViT-L/14-336px~\cite{clip} as the vision encoder and Vicuna-7B~\cite{vicuna} as the LLM. The training details can be found in Sec.~\ref{sec:impl_detail}. 
\subsection{Baseline Methods}
We evaluate TGIF against two complementary categories of baselines: 

\noindent
\textbf{Decoding-based Hallucination Mitigation.}
VCD~\cite{leng2023mitigatingobjecthallucinationslarge}, VTI~\cite{liu2024reducinghallucinationsvisionlanguagemodels}, OPERA~\cite{huang2024opera}, FarSight~\cite{tang2025seeingfarclearlymitigating}, PerturboLLaVA~\cite{chen2025perturbollavareducingmultimodalhallucinations}, SHIELD~\cite{huang2026shield}, and RepEdit-GA~\cite{shi2025exposinghallucinationssuppressthem} intervene at decoding or inference time. TGIF is complementary to these methods: it is a visual feature-level architectural modification trained within the original MLLM pretraining/instruction-tuning pipeline, without requiring an additional hallucination-specific dataset, a separate hallucination-mitigation finetuning stage, or an extra inference-time decoding branch.

\noindent
\textbf{Multi-layer Fusion Baselines.}
DenseConnector~\cite{yao2024denseconnectormllms} and MMFuser~\cite{cao2024mmfusermultimodalmultilayerfeature} aggregate features from multiple vision encoder layers using static fusion strategies.
\subsection{Quantitative Results}
\textbf{Hallucination Benchmarks.}
To assess grounding faithfulness, we adopt two representative hallucination benchmarks: 
HallusionBench (HB)~\cite{hallusionbench} and POPE~\cite{Li-hallucination-2023}. 
HallusionBench probes visual factuality by testing a model’s ability to reject implausible object claims, while POPE reformulates hallucination detection as a binary classification task, quantifying the model’s awareness of object existence.

\noindent
\textbf{OCR Benchmarks.} To evaluate text recognition and detail-sensitive reasoning, we include TextVQA~\cite{singh2019vqamodelsread} and the comprehensive OCRBench~\cite{Liu_2024}, which covers scene-text, document-text, and key information extraction subtasks. These benchmarks reveal the model’s ability to retrieve and interpret fine-grained textual cues—an area where hallucination often manifests due to overreliance on semantic priors.

\noindent
\textbf{Relation Grounding.}
To evaluate fine-grained relational grounding beyond object-existence recognition, we additionally evaluate TGIF on MMRel~\cite{nie2025mmrelbenchmarkingrelationunderstanding}, covering action, spatial, and comparative relations. TGIF improves overall accuracy from 61.0 to 63.0 on 7B and from 64.5 to 75.4 on 13B, with particularly strong gains on spatial-relation subsets (Table~\ref{tab:mmrel_full}). Performance on comparative relations is mixed; the full per-subset breakdown is provided in the Appendix.

\noindent
\textbf{General Reasoning and Overall Benchmarks.} For overall multimodal reasoning performance, we report results on widely used visual QA and instruction-following datasets, including 
MMBench (MMB)~\cite{liu2024mmbench}, 
ScienceQA~\cite{lu2022learn} and 
GQA~\cite{ainslie2023gqatraininggeneralizedmultiquery}. 
Together, these benchmarks measure both the factual grounding and generalization capability of TGIF across visual domains and task types.

\begin{table}[h]
\centering
\label{tab:pope_detailed}
\small
\setlength{\tabcolsep}{3.5pt}
\begin{tabular}{@{}l cc cc cc cc@{}}
\toprule
\multirow{2}{*}{\textbf{Method}} & \multicolumn{2}{c}{\textbf{Adversarial}} & \multicolumn{2}{c}{\textbf{Popular}} & \multicolumn{2}{c}{\textbf{Random}} & \multicolumn{2}{c}{\textbf{Average}} \\
\cmidrule(lr){2-3} \cmidrule(lr){4-5} \cmidrule(lr){6-7} \cmidrule(lr){8-9}
& Acc. & F1 & Acc. & F1 & Acc. & F1 & Acc. & F1 \\
\midrule
\textit{Baseline} & & & & & & & & \\
LLaVA-1.5-7B & 78.96 & 77.57 & 81.88 & 80.06 & 82.29 & 81.33 &  81.04 & 79.65\\
\midrule
\textit{Decoding-Time} & & & & & & & & \\
+ VCD~\cite{leng2023mitigating} & 80.88 & 81.33 & 85.38 & 85.06 & 87.73 & 87.16 & 84.66 & 84.51 \\
+ OPERA~\cite{huang2024opera} & 82.90 & 83.10 & 84.30 & 85.20 & 85.40 & 87.90 & 84.20 & 85.40 \\
+ VTI~\cite{liu2024reducinghallucinationsvisionlanguagemodels} & 84.80 & 84.10 & 86.90 & 86.00 & 87.80 & 87.60 & 86.50 & 85.90 \\
+ FarSight~\cite{tang2025seeingfarclearlymitigating} & 84.40 & 78.50 & 86.20 & 80.10 & 87.70 & 82.60 & 86.10 & 80.40 \\
+ LayerCD~\cite{tong2025mitigatinghallucinationmultimodalllms} & 82.09 & 80.59 & 84.25 & 82.53 & 85.77 & 83.93 & 84.04 & 82.35\\
+ VaLiD~\cite{wang2025validmitigatinghallucinationlarge}& 83.20 & 83.29 & 87.17 & 86.65 & 89.03 & 88.36 & 86.47 & 86.10\\
+ RepEdit-GA~\cite{shi2025exposinghallucinationssuppressthem}
& 81.76 & 82.84
& 86.00 & 86.14
& 89.35 & 89.43
& 85.70 & 86.14 \\
+ SHIELD~\cite{huang2026shield} & 82.50 & 83.60 & 87.40 & 87.60 & \textbf{91.30} & \textbf{91.10} & 87.00 & 87.40\\

\midrule
\textit{Architectural (Ours)} & & & & & & & & \\
\textbf{+ TGIF (Final)} & \textbf{86.00} & \textbf{85.60} & \textbf{88.23} & \textbf{87.60} & 89.51 & 89.10 & \textbf{87.91} & \textbf{87.43} \\
\bottomrule
\end{tabular}
\caption{\textbf{Evaluation on POPE Benchmark.} We report Accuracy (Acc.) and F1-score across three sampling settings: \textit{Adversarial}, \textit{Popular}, and \textit{Random}. TGIF consistently outperforms both the LLaVA-1.5 baseline and specialized decoding-time mitigation methods.}
\end{table}
\begin{table}[t]
\label{tab:HallusionBench}
\begin{center}
\begin{small}

\begin{tabular}{lcr}
\toprule
Method & Params & All Acc $\uparrow$ \\
\midrule
LLaVA-1.5        & 13.0B & 46.94 \\
InstructBLIP~\cite{instructblip}     & 8.2B  & 45.26 \\
MiniGPT5~\cite{Zheng2023MiniGPT5IV}         & 8.2B  & 40.30 \\
\midrule
LLaVA-1.5        & 7.0B  & 46.90 \\
+ VCD~\cite{leng2023mitigating}            & 7.0B  & 46.90 \\
+ OPERA\cite{huang2024opera}         & 7.0B  & 47.10 \\
+ PerturboLLaVA\cite{chen2025perturbollavareducingmultimodalhallucinations}  & 7.0B  & 47.60 \\
\midrule
\textbf{+ TGIF (ours)} & 7.0B & \textbf{49.94} \\
\bottomrule
\end{tabular}
\caption{\textbf{HallusionBench comparison.}
We compare LLaVA-1.5-7B+TGIF with similarly sized open-source models (7–13B). We report GPT4-assisted All Accuracy. TGIF attains an All Accuracy of 49.94\%, outperforming LLaVA-1.5 by +3.0\% and exceeding larger 13B-parameter models. Detailed results are provided in the Appendix. }
\end{small}
\end{center}
\vskip -0.1in
\end{table}
\begin{table}[b]
\label{tab:ocrbench}
\vskip -0.1in
\begin{center}
\begin{small}

\setlength{\tabcolsep}{1pt}
\begin{tabular}{lcccccc}
\toprule
Method & Recog. & $VQA^{S}$ & $VQA^{D}$ & KIE & HMER & Final \\
\midrule
LLaVA-1.5-7B & 160 & 117 & 15 & 5 & 0 & 297 \\
\textbf{+TGIF (ours)} & \textbf{162} & \textbf{121} & \textbf{24} & \textbf{6} & 0 & \textbf{313} \\
\bottomrule
\end{tabular}
\caption{\textbf{OCRBench comparison.}
Recog.:Text Recognition; $VQA^{S}$:Scene Text-centric VQA; $VQA^{D}$:Document-oriented VQA; KIE: Key Information Extraction; HMER:Handwritten Math Expression Recognition. }
\end{small}
\end{center}
\vskip -0.1in
\end{table}

\begin{table*}[t]
\centering
\small
\setlength{\tabcolsep}{4pt}
\begin{tabular}{lccc ccc}
\toprule
\multirow{2}{*}{\textbf{Subset}} &
\multicolumn{3}{c}{\textbf{LLaVA-1.5-7B}} &
\multicolumn{3}{c}{\textbf{LLaVA-1.5-13B}} \\
\cmidrule(lr){2-4}
\cmidrule(lr){5-7}
& Base & TGIF & $\Delta$ & Base & TGIF & $\Delta$ \\
\midrule
dall-e\_action       & 67.5 & 68.0 & +0.5 & 72.8 & 71.3 & -1.5 \\
dall-e\_spatial      & 51.4 & 52.9 & +1.5 & 52.3 & 56.3 & +4.0 \\
spec\_comparative    & 73.2 & 74.3 & +1.1 & 77.4 & 78.5 & +1.1 \\
spec\_spatial        & 52.2 & 56.6 & +4.4 & 54.5 & \textbf{68.8} & \textbf{+14.3} \\
vg\_action           & 71.7 & 73.9 & +2.2 & 73.6 & \textbf{82.8} & \textbf{+9.2} \\
vg\_comparative      & 74.2 & 73.7 & -0.5 & 84.5 & 78.4 & -6.1 \\
vg\_spatial          & 51.2 & 53.1 & +1.9 & 56.0 & \textbf{75.7} & \textbf{+19.7} \\
\midrule
\textbf{Overall}     & 61.0 & \textbf{63.0} & \textbf{+2.0}
                     & 64.5 & \textbf{75.4} & \textbf{+10.9} \\
\bottomrule
\end{tabular}
\caption{\textbf{MMRel Discriminative Y/N results.}
We report accuracy (\%) across action, spatial, and comparative relation subsets for LLaVA-1.5-7B and LLaVA-1.5-13B. TGIF improves overall accuracy at both model scales, with particularly strong gains on spatial-relation subsets for the 13B model.}
\label{tab:mmrel_full}
\end{table*}

\begin{table*}[t]
\centering
\small
\begin{tabular}{@{}l c c ccc@{}}
\toprule
\multirow{2}{*}{\textbf{Method}} & \textbf{Hallucination} & \textbf{OCR} & \multicolumn{3}{c}{\textbf{General Reasoning}} \\
\cmidrule(lr){2-2} \cmidrule(lr){3-3} \cmidrule(lr){4-6}
& POPE (Acc. $\uparrow$) & TextVQA $\uparrow$ & SQA $\uparrow$ & GQA $\uparrow$ & MMB $\uparrow$ \\
\midrule
LLaVA-1.5-7B & 81.04 & 58.20 & 66.80 & 62.00 & 64.30 \\
+ DenseConnector & 86.60 & \textbf{59.20} & \underline{69.50} & \textbf{63.80} & \underline{66.80} \\
+ MMFuser & 86.30 & 58.80 & 68.70 & \underline{62.80} & \textbf{67.50} \\
\midrule
\textbf{+ TGIF (Ours)} & \textbf{87.91} & \underline{58.98} & \textbf{70.10} & 62.58 & 66.40 \\
\bottomrule
\end{tabular}
\caption{\textbf{Main Results on Hallucination and General Reasoning.} \textbf{Hallucination:} POPE; \textbf{OCR:} TextVQA; \textbf{Overall:} ScienceQA (SQA), GQA, MMBench (MMB). Best in \textbf{bold}, second best \underline{underlined}.}
\label{tab:performance_tradeoff}
\end{table*}

\subsection{Router Layer Selection Dynamics}
\begin{figure}[h]
    \centering
    \includegraphics[width=0.7\linewidth]{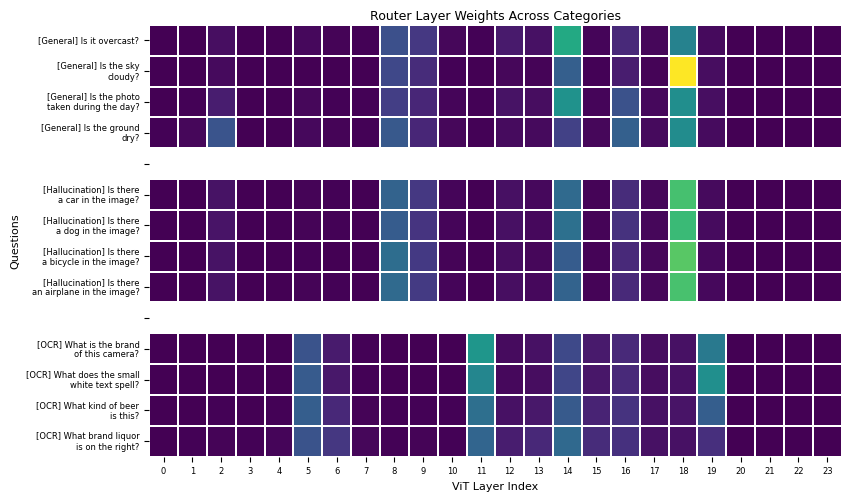}
    \caption{\textbf{Router layer selection patterns across different question categories.}
        This heatmap visualizes the router's learned weights for selecting vision transformer (ViT) layers across three categories of questions: 
        \textbf{General}, \textbf{Hallucination Detection}, and \textbf{OCR/Detail Recognition}. 
        Each row corresponds to one question, and each column indicates a specific ViT layer. 
        Brighter colors denote higher selection weight for that layer. 
        }
    \label{fig:weights}
\end{figure}

To understand how TGIF adapts visual fusion to different query types, we visualize the learned layer-weight distributions in Fig.~\ref{fig:weights}. The router exhibits clear, semantically-driven routing patterns. General queries (e.g., ``Describe the image.'') activate a broad mixture of mid- and high-level layers, reflecting the need for holistic scene understanding. Hallucination-sensitive queries place greater weight on early layers that preserve spatial and boundary cues, which helps verify object presence rather than relying on language priors. In contrast, OCR and detail-oriented questions concentrate weight on mid-to-late layers containing rich text strokes and structural detail. These behaviors confirm that TGIF does not rely on a fixed mixture but instead performs question-aware selection of visual experts, addressing the limitations of static multi-layer fusion.

\subsection{Ablation Studies}

\begin{table*}[t]
\centering
\small
\label{tab:ablation_study}
\setlength{\tabcolsep}{2.5pt} 
\begin{tabular*}{\textwidth}{@{\extracolsep{\fill}}ll cc cc ccc @{}}
\toprule
\multirow{2}{*}{\textbf{Router}} & \multirow{2}{*}{\textbf{Setting}} & \multicolumn{2}{c}{\textbf{Hallucination}} & \multicolumn{2}{c}{\textbf{OCR}} & \multicolumn{3}{c}{\textbf{Reasoning}} \\
\cmidrule(lr){3-4} \cmidrule(lr){5-6} \cmidrule(lr){7-9}
& & POPE $\uparrow$ & HB $\uparrow$ & TVQA $\uparrow$ & OCRB $\uparrow$ & SQA $\uparrow$ & GQA $\uparrow$ & MMB $\uparrow$ \\
\midrule
Text-only & Vanilla ($\lambda{=}0$) & 87.30 & 49.95 & 58.93 & \textbf{31.50} & 68.07 & 62.48 & 65.97 \\
Multi. & Vanilla ($\lambda{=}0$) & 86.26 & \textbf{57.31} & \textbf{59.09} & 29.90 & 69.72 & 62.37 & 65.46 \\
\midrule
\textbf{Text-only} & \textbf{$\lambda{=}0.01$ (Ours)$^\dagger$} & \textbf{87.91} & 49.94 & 58.98 & \underline{31.30} & \textbf{70.10} & \textbf{62.58} & \textbf{66.40} \\
\bottomrule
\end{tabular*}
\caption{\textbf{Ablation Study of TGIF Router Designs.} We compare router architectures and entropy-based load-balancing ($\lambda$). Text-only router with $\lambda{=}0.01$ achieves the best balance across grounding and reasoning. HB: HallusionBench; TVQA: TextVQA; OCRB: OCRBench; SQA: ScienceQA; GQA: GQA; MMB: MMBench.}
\end{table*}
\subsubsection{Text-only vs. Multimodal Routing}
We compare two routing strategies: text-only router that conditions solely on the question embedding, and a multimodal router that additionally incorporates a global image representation.
As shown in Table~\ref{tab:ablation_study}, both variants consistently outperform the LLaVA-1.5 baseline, confirming the effectiveness of dynamic inter-layer fusion.

The multimodal router achieves a substantial gain on HallusionBench, suggesting that global visual context can help the router better distinguish queries requiring semantic reasoning from those demanding conservative verification.
It also improves performance on ScienceQA, indicating stronger generalization on complex, multi-step reasoning tasks.

In contrast, the text-only router performs slightly better on POPE and OCRBench, where hallucination often arises from over-reliance on semantic co-occurrence rather than insufficient visual context. In these cases, conditioning routing solely on query encourages more conservative layer selection, reducing sensitivity to visually plausible but absent objects.
This behavior aligns with our hypothesis that hallucination is strongly influenced by how visual abstraction depth is exposed to the language model.

Overall, while multimodal routing can benefit tasks that require global semantic understanding, we adopt the text-only router as our final configuration due to its stronger and more stable performance on hallucination-sensitive and fine-grained perception benchmarks. This choice also better isolates the effect of query-driven control over visual representation depth, which is the central focus of TGIF.

\subsubsection{Visual Depth Analysis}
To directly test the core hypothesis that different visual depths exhibit different grounding behavior, we construct static variants that expose only early, middle, or late ViT features, as well as a uniform fusion across layers, and evaluate them on POPE-Adversarial. As shown in Table~\ref{tab:pope_adv_depth}, early features are conservative, achieving high precision but low recall, while late features achieve high recall at the cost of precision. No fixed depth is uniformly optimal. TGIF adaptively balances these complementary behaviors and achieves the best F1 among the depth-selection variants.

\begin{table}[t]
\centering
\small
\setlength{\tabcolsep}{4pt}
\begin{tabular}{lcccc}
\toprule
Method & Acc $\uparrow$ & Prec $\uparrow$ & Recall $\uparrow$ & F1 $\uparrow$ \\
\midrule
LLaVA-1.5-7B & 0.790 & 0.814 & 0.741 & 0.776\\
Early-only & 0.804 & \textbf{0.963} & 0.632 & 0.763 \\
Mid-only & 0.840 & 0.942 & 0.725 & 0.819 \\
Late-only & 0.837 & 0.800 & \textbf{0.899} & 0.847 \\
Uniform & 0.838 & 0.957 & 0.707 & 0.813 \\
\textbf{TGIF} & \textbf{0.860} & 0.881 & 0.832 & \textbf{0.856} \\
\bottomrule
\end{tabular}
\caption{\textbf{POPE-Adversarial depth analysis.} Static depth choices exhibit complementary precision--recall trade-offs, while TGIF achieves the best F1 via text-guided inter-layer fusion.}
\label{tab:pope_adv_depth}
\end{table}

\subsubsection{Effect of Load-Balancing Coefficient $\lambda$}
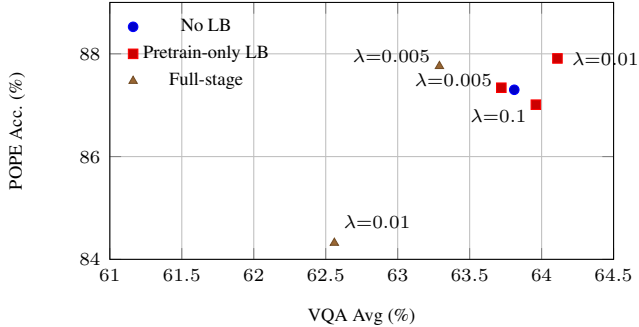
\begin{figure}[t]
\centering
\begin{tikzpicture}
\begin{axis}[
    width=0.6\linewidth,
    height=4.0cm,
    xlabel={VQA Avg (\%)},          
    ylabel={POPE Acc. (\%)},
    xmin=61, xmax=64.5,
    ymin=84, ymax=89,
    grid=both,
    tick label style={font=\scriptsize},
    label style={font=\scriptsize},
    legend style={
        at={(0.02,0.98)},
        anchor=north west,
        draw=none,
        fill=none,
        font=\scriptsize
    },
    clip=false 
]

\addplot+[
    only marks,
    mark=*,
    mark size=1.8pt
] coordinates {
    (63.81, 87.30)  
};
\addlegendentry{No LB}

\addplot+[
    only marks,
    mark=square*,
    mark size=1.8pt
] coordinates {
    (64.11, 87.91)  
    (63.72, 87.34)  
    (63.96, 87.01)  
};
\addlegendentry{Pretrain-only LB}

\node[anchor=west, font=\scriptsize, xshift=2pt] 
    at (axis cs:64.11,87.91) {$\lambda{=}0.01$};
\node[anchor=south east, font=\scriptsize, yshift=-3pt] 
    at (axis cs:63.72,87.34) {$\lambda{=}0.005$};
\node[anchor=north east, font=\scriptsize, yshift=2pt] 
    at (axis cs:63.96,87.01) {$\lambda{=}0.1$};

\addplot+[
    only marks,
    mark=triangle*,
    mark size=1.8pt
] coordinates {
    (62.56, 84.32)  
    (63.29, 87.76)  
};
\addlegendentry{Full-stage}

\node[anchor=south west, font=\scriptsize, yshift=2pt] 
    at (axis cs:62.56,84.32) {$\lambda{=}0.01$};
\node[anchor=south east, font=\scriptsize, yshift=-2pt] 
    at (axis cs:63.29,87.76) {$\lambda{=}0.005$};

\end{axis}
\end{tikzpicture}
\caption{\textbf{Effect of load balancing on the VQA–hallucination trade-off.}
Each point shows the average VQA score (ScienceQA, GQA, TextVQA) versus POPE accuracy
for a different router / load-balancing configuration. We annotate all load-balancing
settings next to their corresponding points.}
\label{fig:lambda_tradeoff}
\end{figure}
\begin{figure*}[t]
    \centering
    \includegraphics[width=0.8\linewidth, trim={0cm 4cm 0cm 6cm},clip]{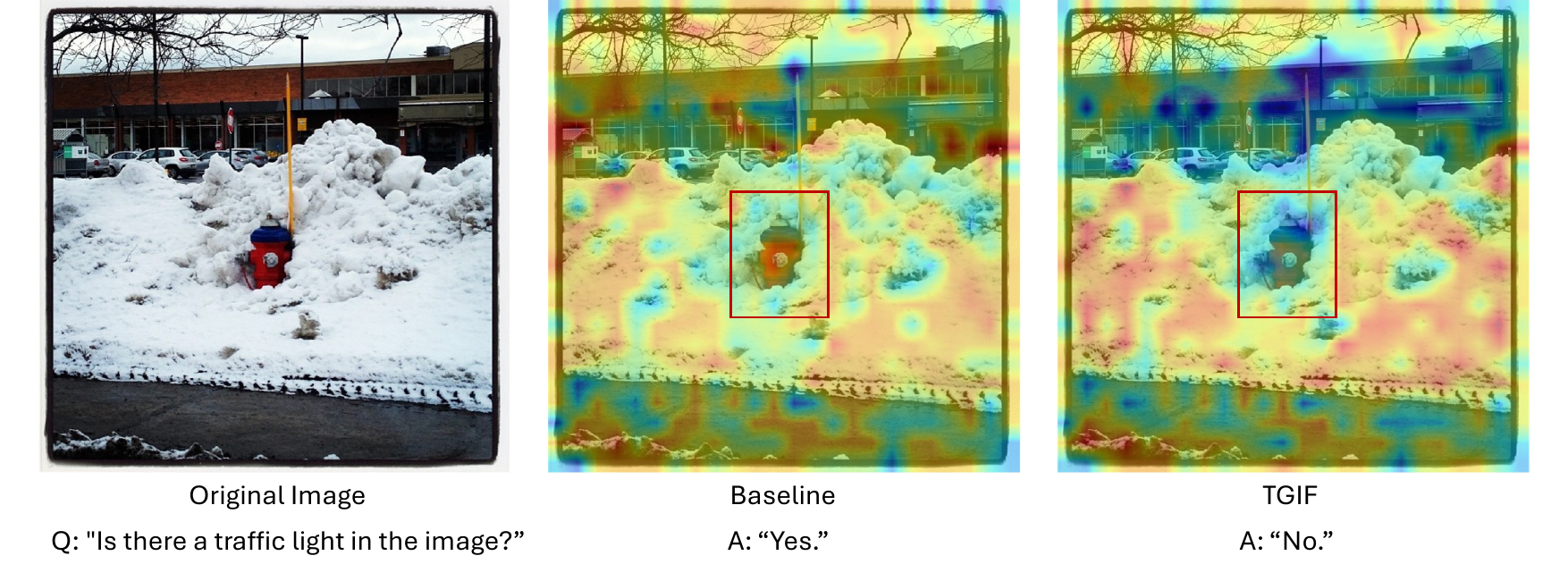}
    \caption{\textbf{Qualitative comparison on POPE adversarial.}
    \textbf{Left:} original image.
    \textbf{Middle:} baseline LLaVA text-conditioned relevance map.
    \textbf{Right:} TGIF text-conditioned relevance map.
    The baseline model attends strongly to a semantically misleading object (a fire hydrant resembling a traffic light), resulting in a hallucinated positive answer.}
    \label{fig:qual_fix}
\end{figure*}

We further analyze the impact of the entropy-based load-balancing loss on routing stability and downstream performance. As shown in Fig.~\ref{fig:lambda_tradeoff}, applying a small amount of regularization during pretraining only provides the best balance between VQA accuracy and hallucination robustness. In particular, the $\lambda{=}0.01$ pretrain-only configuration yields both the highest POPE accuracy and the strongest average VQA score, indicating that mild early-stage entropy encourages the router to explore a diverse set of layers without suppressing its ability to specialize.

In contrast, larger coefficients (e.g., $\lambda{=}0.1$) or applying the loss throughout full fine-tuning tend to over-regularize the router, pulling the layer distribution toward uniformity and reducing its ability to adapt the fusion pattern to the input query. These settings achieve weaker VQA performance and, in some cases, degraded hallucination resistance. Overall, the results suggest that light, pretraining-only regularization is crucial: it stabilizes routing and prevents expert collapse while still allowing the model to learn query-dependent, discriminative layer selection during instruction tuning.

\subsection{Qualitative comparison}

Figure~\ref{fig:qual_fix} illustrates how text-guided inter-layer fusion alters the visual evidence exposed to the language model.
The adversarial query asks whether a traffic light is present.
Although the image contains no traffic light, the baseline LLaVA attends strongly to a centrally located fire hydrant whose color and shape are highly correlated with traffic lights in the training distribution.
This over-reliance on late, semantically aligned visual features leads the model to hallucinate a positive answer.

In contrast, TGIF dynamically reweights visual representations across layers conditioned on the text query.
The resulting relevance map deemphasizes the misleading object and distributes attention more broadly across the scene, suppressing language-driven semantic shortcuts.
As a result, TGIF correctly answers “No.”
While this is a single example, it qualitatively demonstrates how text-guided layer fusion modulates the balance between visual abstraction and language priors, consistent with our quantitative improvements on hallucination benchmarks.

\subsection{Computation and Memory Overhead}
We analyze the parameter, latency, and memory overhead introduced by TGIF to validate its lightweight design.
The minimal runtime overhead is expected for two reasons. First, the router is a shallow MLP evaluated once per input and contributes negligible computation compared to the frozen vision encoder and the large language model. Second, LLaVA already computes and stores intermediate visual features across layers during the forward pass, so TGIF does not introduce additional vision encoder computation or memory allocation. 

\begin{table}[h]
\label{tab:overhead}
\begin{center}
\begin{small}

\setlength{\tabcolsep}{3pt}
\renewcommand{\arraystretch}{1}
\begin{tabular}{lccc}
\toprule
Model & Params (M) & Latency (s) & Peak Mem (GB) \\
\midrule
LLaVA      & 7062.9 & 0.2203 & 27.16 \\
\textbf{+ TGIF}     & 7065.0 & 0.2224 & 27.16 \\
\midrule
Overhead   & +0.03\% & +0.93\% & +0.00 \\
\bottomrule
\end{tabular}
\caption{\textbf{Computation and memory overhead.}
TGIF introduces negligible overhead relative to the LLaVA-1.5 baseline.}
\end{small}
\end{center}
\vskip -0.2in
\end{table}
\section{Conclusion}
We present TGIF, a lightweight text-guided inter-layer fusion framework that dynamically controls which visual representations are exposed to the language model. By adapting the visual abstraction level to the input query, TGIF improves visual grounding and reduces hallucination without modifying the vision encoder, increasing token count, or introducing significant computational overhead. Extensive experiments across hallucination, OCR, and general VQA benchmarks demonstrate that TGIF consistently outperforms strong baselines while preserving overall reasoning capability. These results suggest that query-adaptive control of visual feature depth is a promising direction for building more reliable and trustworthy multimodal language models.

\bibliography{colm2026_conference}
\bibliographystyle{colm2026_conference}

\newpage
\appendix
\section{Appendix}

\section{Architectural Comparison with Prior Layer Fusion Designs}
\label{sec:arch_comparison}
\begin{figure*}[h]
    \centering
    \includegraphics[width=\textwidth,trim={0cm 3cm 0cm 6.5cm},clip]{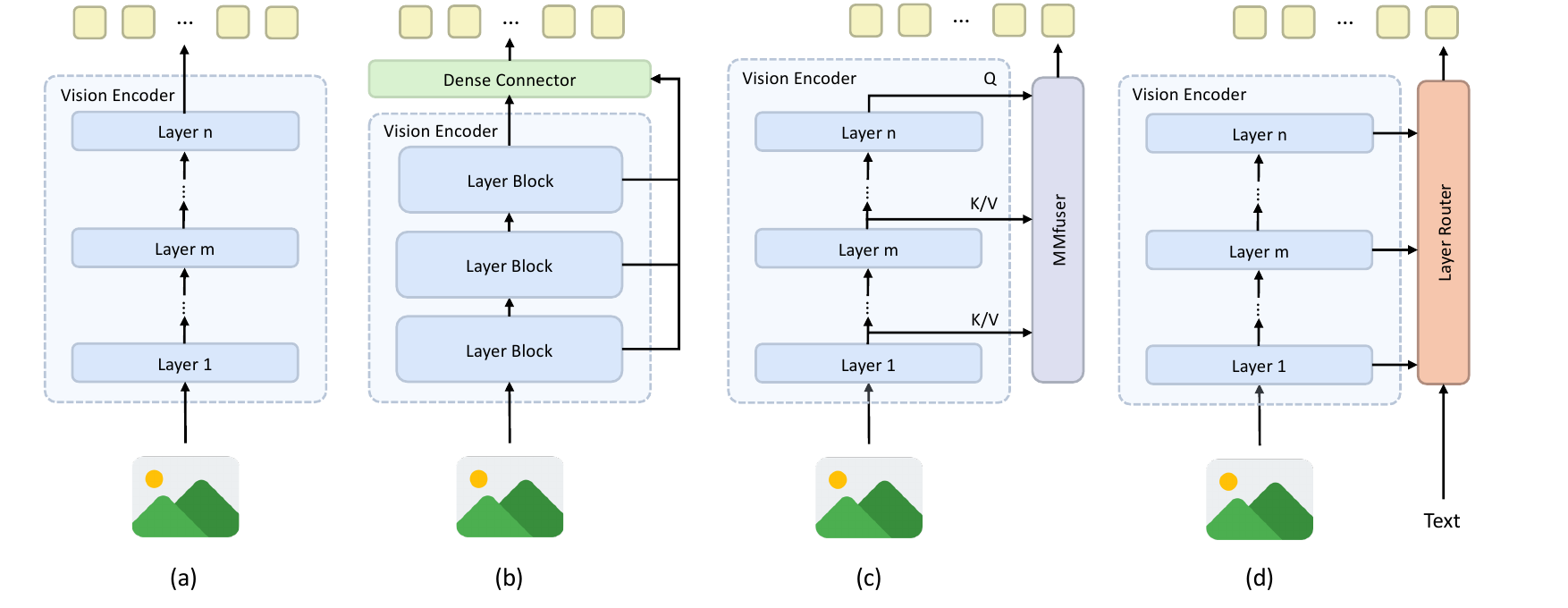}
    \caption{\textbf{Comparison of layer fusion designs in MLLMs.}
    (a) \textbf{MLP Connector}: Projects visual tokens from a single, fixed (typically penultimate) vision encoder layer.
    (b) \textbf{Dense Connector}: Aggregates features from multiple layers via concatenation or downsampling with a static fusion pattern.
    (c) \textbf{MMFuser}: Retrieves shallow-layer features using deep-layer queries through cross-attention to recover local details.
    (d) \textbf{TGIF (ours)}: Introduces a text-guided router that dynamically reweights visual features across layers based on the input query, enabling adaptive control over visual abstraction depth.
    }
    \label{fig:comparison}
\end{figure*}

Figure~\ref{fig:comparison} provides a schematic comparison between commonly used multimodal connectors and the proposed text-guided inter-layer fusion (TGIF).

\textbf{Single-layer projection.}
Most existing MLLMs, including LLaVA-style architectures, employ a simple MLP connector that projects visual tokens from a single, fixed layer of the vision encoder (typically the penultimate layer) into the LLM.
While computationally efficient, this design exposes only late-stage semantic representations to the language model, limiting access to low-level and mid-level visual evidence that is often crucial for fine-grained grounding and hallucination verification.

\textbf{Static multi-layer fusion.}
Recent works enrich visual representations by aggregating features from multiple encoder layers.
DenseConnector concatenates or downsamples features from selected depths, while MMFuser retrieves shallow-layer details using deep-layer queries via cross-attention.
Although these approaches improve visual richness, the fusion strategy remains \emph{static}: the set of layers and their relative contributions are fixed once chosen and do not adapt to the input query.

\textbf{Text-guided inter-layer fusion (TGIF).}
In contrast, TGIF treats the layers of a frozen vision encoder as a pool of specialized visual experts and dynamically selects among them conditioned on the input text.
A lightweight router predicts a soft distribution over layers for each query, allowing the model to adaptively control which level of visual abstraction is exposed to the LLM.
This enables conservative, low-level verification for hallucination-sensitive queries and higher-level semantic aggregation for descriptive or reasoning-oriented prompts.

Overall, this comparison highlights a key distinction between TGIF and prior fusion methods:
rather than statically enriching visual representations, TGIF introduces query-dependent control over visual abstraction depth, directly addressing hallucination arising from mismatched visual evidence.

\section{Router Variants: Text-only vs.\ Multimodal}
\label{sec:router_variants}

\begin{figure}[t]
    \centering
    \includegraphics[width=0.8\linewidth,trim={0cm 4cm 12cm 4cm},
  clip]{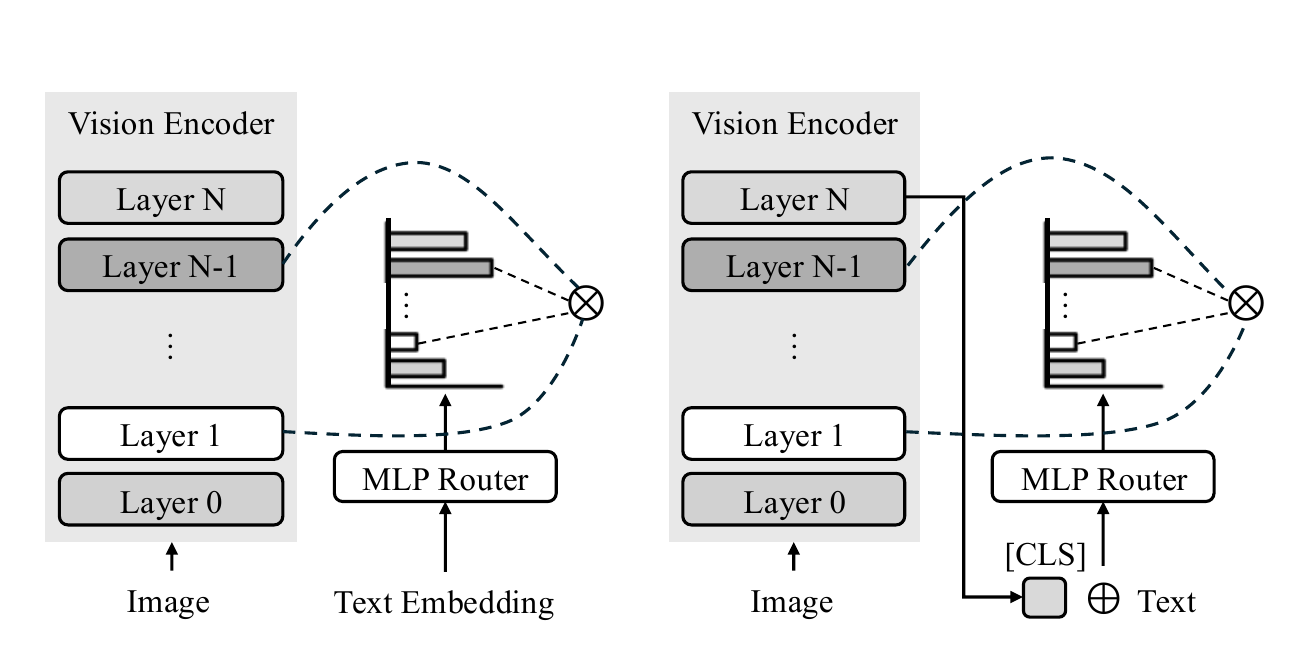}
    \caption{\textbf{Text-only vs.\ multimodal router in TGIF.}
    \textbf{Left:} The text-only router takes the question embedding $\mathbf{f}_{\text{text}}$ and outputs layer weights $w_{\text{layer}}$ via an MLP and softmax.
    \textbf{Right:} The multimodal router concatenates a projected text embedding $\mathbf{f}_{\text{text}}\mathbf{W}_t$ and a projected global image embedding $\mathbf{f}_{\text{image}}\mathbf{W}_v$ (penultimate-layer [CLS]) to form $\mathbf{f}_{\text{multi}}\in\mathbb{R}^{2D_p}$ before predicting $w_{\text{layer}}$.
    }
    \label{fig:router_variants}
\end{figure}

Figure~\ref{fig:router_variants} illustrates the difference between the two routing variants used in TGIF.
The \textbf{text-only router} predicts layer weights solely from the question embedding, while the \textbf{multimodal router} additionally incorporates a global image representation (the penultimate-layer [CLS] token), enabling routing to depend on both query intent and coarse image content.

\section{Training Details}
\label{sec:impl_detail}

All experiments are conducted on a single node equipped with $8\times$ NVIDIA H100 (80GB) GPUs. We utilize DeepSpeed ZeRO (Stage 2 for pretraining, Stage 3 for instruction tuning) to optimize memory efficiency. Unless otherwise specified, we employ Vicuna-7B-v1.5 as the language model and CLIP ViT-L/14@336px as the frozen vision encoder. Training is performed with \texttt{bf16} precision and \texttt{tf32} matrix multiplication acceleration.

\subsection{Training Recipe}
All models are trained following the two-stage training paradigm of LLaVA-1.5~\cite{li2023llava}, consisting of
\textbf{Stage~1: Feature Alignment Pretraining} and
\textbf{Stage~2: Instruction Finetuning}.
For a fair comparison, we adopt the same training datasets and data splits as LLaVA-1.5.

\vspace{4pt}\noindent\textbf{Stage 1: Feature Alignment.}
Following the LLaVA-1.5 protocol, we train only the multimodal projector (including the proposed TGIF router) for 1 epoch on the LLaVA pretraining dataset (558K image–text pairs). We utilize a learning rate of $1\times10^{-3}$ with a cosine decay schedule and a warmup ratio of 0.03. The global batch size is set to 32 (gradient accumulation steps $= 1$), with a maximum sequence length of 2048. Weight decay is set to 0.

\vspace{4pt}\noindent\textbf{Stage 2: Instruction Tuning.}
In the second stage, we fine-tune the projector and the LLM on the LLaVA v1.5 mixture (665K samples) for 1 epoch. We reduce the learning rate to $2\times10^{-5}$ while maintaining the cosine schedule, warmup ratio of 0.03, and weight decay of 0. The per-GPU batch size is adjusted to 16.

Throughout both stages, the vision encoder remains entirely frozen. We enable gradient checkpointing to conserve memory. All specific hyperparameters and script templates are provided in the attached code for reproducibility.

\section{Benchmark Descriptions}

\subsection{Hallucination Evaluation}

\paragraph{POPE~\cite{Li-hallucination-2023}.}
POPE assesses object hallucination via a binary verification task. For a given image, the model must answer Yes/No questions (e.g., ``Is there a \textless object\textgreater{} in the image?''). Negative samples are generated using three strategies: (1)~\textbf{Random} sampling, (2)~\textbf{Popular} COCO categories, and (3)~\textbf{Adversarial} co-occurring objects. This setup isolates visual grounding capabilities from captioning priors. We report Accuracy, Precision, Recall, F1-score, and the "Yes" response ratio.

\paragraph{HallusionBench~\cite{hallusionbench}}
This benchmark evaluates visual factual grounding by presenting questions with fabricated or contradictory premises. Unlike standard VQA, HallusionBench penalizes models for failing to reject incorrect visual claims regarding object existence, attributes, and spatial relations. Following the official protocol, we employ a GPT-4 assisted evaluation metric. We report \textbf{aAcc} (All Accuracy) as the primary metric, alongside \textbf{qAcc} (Question-Pair Accuracy) and \textbf{fAcc} (Figure Accuracy).

\subsection{OCR Evaluation}

\paragraph{OCRBench~\cite{Liu_2024}.}
OCRBench is a comprehensive evaluation suite consisting of 1,000 manually verified QA pairs aggregated from 29 diverse datasets. It assesses five core capabilities: Text Recognition, Scene Text VQA, Document VQA, Key Information Extraction (KIE), and Handwritten Math Expression Recognition (HMER). Evaluation is performed via exact string matching.

\paragraph{TextVQA~\cite{Singh_2019_CVPR}.}
TextVQA is a benchmark designed to evaluate a model’s ability to read and reason about text embedded in natural images. It contains over 45K questions paired with images rich in scene text (e.g., signs, posters, menus), where answering explicitly requires OCR and text-based reasoning. The dataset features a highly diverse, open-ended answer space, making it a challenging testbed for fine-grained visual grounding and text-centric multimodal understanding.

\subsection{General Reasoning and Overall Benchmarks.}
\paragraph{ScienceQA~\cite{lu2022learn}.} ScienceQA is a large-scale multimodal multiple-choice benchmark designed to evaluate multi-hop scientific reasoning across vision and language. It contains over 21K questions drawn from elementary and high school science curricula, spanning natural, social, and language sciences. Nearly half of the questions include visual context, and most are annotated with grounded lectures and detailed explanations, enabling analysis of both answer accuracy and reasoning quality. ScienceQA is widely used to assess general multimodal reasoning and instruction-following capabilities beyond surface-level perception.

\paragraph{GQA~\cite{hudson2019gqa}.}
GQA is a large-scale visual question answering benchmark designed to evaluate compositional and multi-step visual reasoning. It contains over 22M questions grounded in real-world images, each paired with a scene graph encoding objects, attributes, and relations. Questions are generated to require explicit reasoning over these structured representations, reducing language priors and dataset bias. In addition to accuracy, GQA provides fine-grained metrics to assess consistency and plausibility, making it a standard benchmark for general visual reasoning and scene understanding.

\paragraph{MMBench~\cite{liu2024mmbench}} MMBench is a large-scale, fine-grained benchmark designed to systematically evaluate the perceptual and reasoning abilities of vision–language models. It contains 2,974 multiple-choice questions spanning 20 ability dimensions, organized into a three-level taxonomy covering coarse perception, fine-grained perception, and multimodal reasoning. Instead of task-specific evaluation, MMBench assesses models across diverse capability dimensions, providing more diagnostic feedback. To ensure robust and reproducible evaluation, model outputs are matched to answer choices using an LLM-based evaluator, enabling reliable comparison across open-ended generation models.

\section{Additional Quantitative Results}

\subsection{HallusionBench Leaderboard}
Table~\ref{tab:hallusion_leaderboard} presents the full HallusionBench correctness leaderboard. TGIF achieves competitive performance among 7B parameters models, particularly in the aAcc metric, surpassing the LLaVA-1.5 baseline and approaching the performance of proprietary models like Gemini Pro Vision. 

Notably, despite utilizing a significantly smaller language backbone (7B parameters), TGIF (49.94\%) outperforms several larger open-source baselines, including the 13B-parameter LLaVA-1.5 (46.94\%) and the 12.1B-parameter BLIP2-T5 (48.09\%)~\cite{liu2023improved,Li2023BLIP2BL}. It secures the third rank overall, trailing only the closed-source models GPT-4V and Claude 3~\cite{2023GPT4VisionSC,claude3}. This result shows the efficiency of our text-guided fusion strategy: by dynamically routing to the most relevant visual features, TGIF extracts rich grounding signals from the frozen encoder, effectively mitigating hallucination even against models with nearly double the parameter count.

\begin{table*}[t]
\caption{\textbf{HallusionBench correctness leaderboard.}
We report Question-pair Accuracy (\textit{qAcc}), Figure Accuracy (\textit{fAcc}),
and All Accuracy (\textit{aAcc}). Top-3 models under GPT-4--assisted evaluation
are highlighted in \textbf{bold}.}
\label{tab:hallusion_leaderboard}
\begin{center}
\begin{small}

\setlength{\tabcolsep}{6pt}
\begin{tabular}{l c c c c c}
\toprule
Method & Params & Eval & qAcc $\uparrow$ & fAcc $\uparrow$ & aAcc $\uparrow$ \\
\midrule
\multirow{2}{*}{GPT-4V~\cite{2023GPT4VisionSC}} 
& \multirow{2}{*}{--} & Human & 31.42 & 44.22 & 67.58 \\
&  & GPT-4 & \textbf{28.79} & \textbf{39.88} & \textbf{65.28} \\
\midrule
\multirow{2}{*}{LLaVA-1.5~\cite{liu2023improved}} 
& \multirow{2}{*}{13B} & Human & 9.45 & 25.43 & 47.12 \\
&  & GPT-4 & 10.55 & \textbf{24.86} & 46.94 \\
\midrule
Claude~3~\cite{claude3} & -- & GPT-4 & \textbf{21.76} & \textbf{28.61} & \textbf{56.86} \\
Gemini Pro Vision~\cite{google_gemini} & -- & GPT-4 & 7.69 & 8.67 & 36.85 \\
\midrule
BLIP2-T5~\cite{Li2023BLIP2BL} & 12.1B & GPT-4 & 15.16 & 20.52 & 48.09 \\
Qwen-VL~\cite{Bai2023QwenVLAF} & 9.6B & GPT-4 & 5.93 & 6.65 & 39.15 \\
Open-Flamingo~\cite{alayrac2022flamingo} & 9B & GPT-4 & 6.37 & 11.27 & 38.44 \\
MiniGPT-5~\cite{Zheng2023MiniGPT5IV} & 8.2B & GPT-4 & 10.55 & 9.83 & 40.30 \\
MiniGPT-4~\cite{zhu2023minigpt} & 8.2B & GPT-4 & 8.79 & 10.12 & 35.78 \\
InstructBLIP~\cite{dai2023instructblip} & 8.2B & GPT-4 & 9.45 & 10.11 & 45.26 \\
BLIP-2~\cite{Li2023BLIP2BL} & 8.2B & GPT-4 & 5.05 & 12.43 & 40.48 \\
mPLUG-Owl v2~\cite{ye2023mplugowl2} & 8.2B & GPT-4 & 13.85 & 19.94 & 47.30 \\
mPLUG-Owl v1~\cite{ye2023mplug} & 7.2B & GPT-4 & 9.45 & 10.40 & 43.93 \\
LRV-Instruction~\cite{liu2023aligning} & 7.2B & GPT-4 & 8.79 & 13.01 & 42.78 \\
\midrule
\textbf{LLaVA-1.5 + TGIF (Ours)} & 7B & GPT-4 & \textbf{17.36} & 23.70 & \textbf{49.94} \\
\midrule
GIT~\cite{Wang2022GITAG} & 0.8B & GPT-4 & 5.27 & 6.36 & 34.37 \\
\midrule
Random Chance & -- & GPT-4 & 15.60 & 18.21 & 45.96 \\
\bottomrule
\end{tabular}

\end{small}
\end{center}
\vskip -0.1in
\end{table*}

\subsection{OCRBench Performance}
We report detailed OCRBench scores in Table~\ref{tab:OCRBench}. TGIF demonstrates improvements in Scene Text VQA ($VQA^S$) and Document VQA ($VQA^D$), contributing to a higher final score compared to the LLaVA-1.5 baseline. This suggests that layer fusion effectively captures fine-grained textual details often lost in single-layer embeddings.

\begin{table*}[h]
\caption{\textbf{Detailed results on OCRBench.}
Breakdown by sub-task: Text Recognition (Recog.),
Scene Text VQA ($VQA^{S}$),
Document VQA ($VQA^{D}$),
Key Information Extraction (KIE),
and Handwritten Math Expression Recognition (HMER).
Best results are shown in \textbf{bold}.}
\label{tab:OCRBench}
\begin{center}
\begin{small}

\setlength{\tabcolsep}{6pt}
\begin{tabular}{l c c c c c c}
\toprule
Method & Recog. & $VQA^{S}$ & $VQA^{D}$ & KIE & HMER & Total \\
\midrule
Gemini Pro~\cite{google_gemini} 
& \textbf{215} & \textbf{174} & 128 & 134 & 8 & \textbf{659} \\
GPT-4V~\cite{2023GPT4VisionSC} 
& 167 & 163 & \textbf{146} & \textbf{160} & \textbf{9} & 645 \\
\midrule
mPLUG-Owl2~\cite{ye2023mplugowl2} 
& 153 & 153 & 41 & 19 & 0 & 366 \\
LLaVAR~\cite{Zhang2023LLaVAREV} 
& 186 & 122 & 25 & 13 & 0 & 346 \\
LLaVA-1.5-13B~\cite{liu2023improved} 
& 176 & 129 & 19 & 7 & 0 & 331 \\
\midrule
LLaVA-1.5-7B~\cite{liu2023improved} 
& 160 & 117 & 15 & 5 & 0 & 297 \\
\textbf{+TGIF (Ours)} 
& 162 & 121 & 24 & 6 & 0 & 313 \\
\midrule
mPLUG-Owl~\cite{mplug-owl} 
& 172 & 104 & 18 & 3 & 0 & 297 \\
InstructBLIP~\cite{instructblip} 
& 168 & 93 & 14 & 1 & 0 & 276 \\
BLIP-2~\cite{Li2023BLIP2BL} 
& 154 & 71 & 10 & 0 & 0 & 235 \\
MiniGPT-4 v2~\cite{minigpt-4} 
& 124 & 29 & 4 & 0 & 0 & 157 \\
\bottomrule
\end{tabular}

\end{small}
\end{center}
\vskip -0.1in
\end{table*}

\end{document}